\documentclass[preprint,epsfig,onecolumn,floats, showkeys]{revtex4}
\usepackage{makeidx}
\usepackage{amsmath}
\usepackage{amssymb}
\usepackage{graphicx}
\usepackage{epsfig}

\begin{document}

\title{Experiment Study of Entropy Convergence of Ant Colony Optimization}
\author{Chao-Yang Pang}
\email{cypang@live.com}
\email{cypang@sicnu.edu.cn}
\affiliation{Group of Gene Computation, Key Lab. of Visual Computation and Virtual
Reality, Sichuan Province,Chengdu 610068, China;\\
Group of Gene Computation,College of Mathematics and Software Science,
Sichuan Normal University, Chengdu 610066, China}
\author{Chong-Bao Wang}
\affiliation{Group of Gene Computation,College of Mathematics and Software Science,
Sichuan Normal University,}
\affiliation{Chengdu 610066, China}
\author{Ben-Qiong Hu}
\affiliation{College of Information Management, Chengdu University of Technology, 610059,
China}

\begin{abstract}
Ant colony optimization (ACO) has been applied to the field of combinatorial
optimization widely. But the study of convergence theory of ACO is rare
under general condition. In this paper, the authors try to find the evidence
to prove that entropy is related to the convergence of ACO, especially to
the estimation of the minimum iteration number of convergence. Entropy is a
new view point possibly to studying the ACO convergence under general
condition.
\end{abstract}

\maketitle


\section{Introduction}

ACO is a recently developed, population-based approach presented by M.
Dorigo and A. Colorni etc. al., it was inspired by the ants' foraging
behavior in 1991 \cite{Dorigo,Colorni,Dorigo1}. Ant System (AS) was first
introduced in three different versions \cite{Dorigo,Colorni,Dorigo1}, they
were called ant-density, ant-quantity, and ant-cycle. Ant Colony System
(ACS) has been introduced in \cite{Dorigo2,Cambardella} to improve the
performance of AS. Later, AS and ACS developed into a unifying framework to
solve combinatorial optimization problems \cite{Dorigo3,Dorigo4}, \ and the
framework is often called Ant Colony Optimization (ACO). ACO has been
applied to solve optimization problems\cite{Ball1,Ball2}, such as Traveling
Salesman Problem (TSP)\cite{Dorigo2,Cambardella1}, Quadratic Assignment
Problem(QAP)\cite{Cambardella2}, Job-shop Scheduling Problem(JSP)\cite%
{Cambardella1}, Vehicle Routing Problem( VRP)\cite{Bullnheimer,Forsyth} and
Data Mining(DM)\cite{Rafael}. The high performance of ACO and its wide
application make it as famous as other optimization algorithms, such as
Simulated Annealing (SA)\cite{Kirkpatrick}, Tabu Search (TS)\cite{Glover},
Genetic Algorithms (GA)\cite{Golderg}, and so on.

The study of ACO theory is necessary but rare. W. J. Gutjahr studies the
convergence of ACO under some conditions by Graph Theory\cite{Gutjahr},
which is called Graph-Based Ant System (GBAS). GBAS maps a feasible solution
of optimization problem to a route in a directed graph. T. St$\overset{..}{u}
$ezle and M. Dorigo proved the existence of the ACO convergence under two
conditions, one is to only update the pheromone of the shortest route
generated at each iteration step, the other is that the pheromone on all
routes has lower bound \cite{Stuezle}. J. H. Yoo analyzes the convergence of
a kind of distributed ants routing algorithm by the method of artificial
intelligence \cite{Yoo,Yoo1}. Sun analyzes the convergence of a simple ant
algorithm by Markov Process\cite{Sun}. Ding presents a hybrid algorithm of
ACO and genetic algorithm, and analyzes the convergence by Markov theory
\cite{Ding}. Hou presents a special ACO algorithm and proves its convergence
by fixed-point theorem \cite{Hou}.

The ways of studying ACO convergence are rare, such as Markov theory, Graph
Theory, and so on. And only the results with some constraint conditions are
obtained currently, and the result with no constraint condition is still
unknown. The motivation of this paper is to explore the way to study ACO
convergence under no constraint condition.

\section{Framework of ACO}

In the 1990s, ACO was introduced as a novel nature-inspired method for the
solution of hard combinatorial optimization problems (Dorigo, 1992; Dorigo
et al., 1996, 1999; Dorigo and St$\overset{..}{u}$ezle, 2004). The inspiring
source of ACO is the foraging behavior of real ants. When searching for
food, ants initially explore the area surrounding their nest in a random
manner. As soon as an ant finds a food source, it remembers the route passed
by and carries some food back to the nest. During the return trip, the ant
deposits pheromone on the ground. The deposited pheromone, guides other ants
to the food source. And it has been shown (Goss et al., 1989), indirect
communication among ants via pheromone trails enables them to find the
shortest routes between their nest and food sources.

The framework of ACO is shown in \textbf{Algorithm 1}, and it is applied to
solve Travel Salesman Problem (TSP). Where TSP can be explained as follows:
for a given set of cities, the task of TSP is to find the cheapest route of
visiting all of the cities and returning to starting point, provided each
city is only visited once.

\bigskip \textbf{Algorithm 1}

\textbf{Step1.} Initialization: Initialize pheromone of all edges among
cities. And put $m$ ants at different cities randomly. Pre-assign an
iteration number $N_{C_{\max }}$ and let $t=0$, where $t$ denotes the $t-th$
iteration step.

\bigskip \textbf{Step2. }while($t<N_{C_{\max }}$)

\{

\textbf{Step2.1}. All ants select its next city according to the transition
probability defined in formula (1), which is the probability that the $k-th$
ant selecting the edge from $i-th$ city to $j-th$ city.

\bigskip
\begin{equation}
p_{ij}^{(k)}(t)=\left\{
\begin{tabular}{ccc}
$\frac{\tau _{ij}^{\alpha }(t).\eta _{ij}^{\beta }}{\underset{s\in
allowed_{k}}{\sum }\tau _{is}^{\alpha }(t).\eta _{is}^{\beta }}$ & $if$ & $%
j\in allowed_{k}$ \\
$0$ & \multicolumn{2}{c}{$otherwise$}%
\end{tabular}%
\right.  \label{eq1}
\end{equation}%
, where $allowed_{k}$ denotes the set of cities that can be accessed by the $%
k-th$ ant; $\tau _{ij}(t)$ is the pheromone value of the edge ($i,j$); $\eta
_{ij}$\ is a local heuristic function defined as

\bigskip
\begin{equation}
\eta _{ij}=\frac{1}{d_{ij}}  \label{eq2}
\end{equation}%
, where $d_{ij}$ is the distance between the $i-th$\ city and the $j-th$
city; the parameters $\alpha $ and $\beta $\ determine the relative
influence of the trail strength and the heuristic information respectively.

\ \textbf{Step2.2.} After all ants finish their travels, all pheromone
values $\tau _{ij}(t)$ are updated according to formula (\ref{eq3}).

\bigskip
\begin{equation}
\tau _{ij}(t+1)=(1-\rho )\cdot \tau _{ij}(t)+\Delta \tau _{ij}(t)
\label{eq3}
\end{equation}%
\begin{equation}
\Delta \tau _{ij}(t)=\overset{m}{\underset{k=1}{\sum }}\Delta \tau
_{ij}^{(k)}(t)  \label{eq4}
\end{equation}

\bigskip
\begin{equation}
\Delta \tau _{ij}^{(k)}(t)=\left\{
\begin{tabular}{ccc}
$\frac{Q}{L^{(k)}(t)}$ & $if$ & $the$ $k-th$ $ant$ $pass$ $edge$ $(i,j)$ \\
$0$ & \multicolumn{2}{c}{$otherwise$}%
\end{tabular}%
\right.   \label{eq5}
\end{equation}%
,where $L^{(k)}(t)$ is the length of the route passed by the $k-th$ ant; $%
\rho $ is the persistence ratio of the trail (thus, ($1-\rho $)\ corresponds
to the evaporation ratio); $Q$ denotes constant quantity of pheromone.

\textbf{Step2.3.} Increase iteration number, i.e., $t\leftarrow t+1$.

\}

\textbf{Step3.} End procedure and select the route which has shortest length
as the output.

\section{The Statistical Feature of The Solutions of ACO}

\subsection{Definition of Symbols}

ACO solving the problem of TSP is the model in this paper. Suppose the $m$
ants are $a_{1}$, $a_{2}$, ......, $a_{m}$. At the $t-th$\ iteration step,
ant $a_{i}$ selects route $r_{i}^{(t)}$ and it has length $L_{i}^{(t)}$.
After all ants finish their $t-th$ travels,$\ $there are totally amount of
pheromone $f_{i}^{(t)}=\underset{(k,j)\in r_{i}^{(t)}}{\sum \tau _{kj}(t)}$
depositing on router $r_{i}^{(t)}$, where $\tau _{kj}(t)$ denotes pheromone
depositing at the edge $(k,j)$ by all ants.

\textbf{Definition 1} \textbf{(Pheromone Probability)}:

\begin{equation}
p_{i}^{(t)}=\frac{f_{i}^{(t)}}{\overset{m}{\underset{j=1}{\sum }}f_{j}^{(t)}}
\label{eq6}
\end{equation}

\begin{figure}[tbh]
\epsfig{file=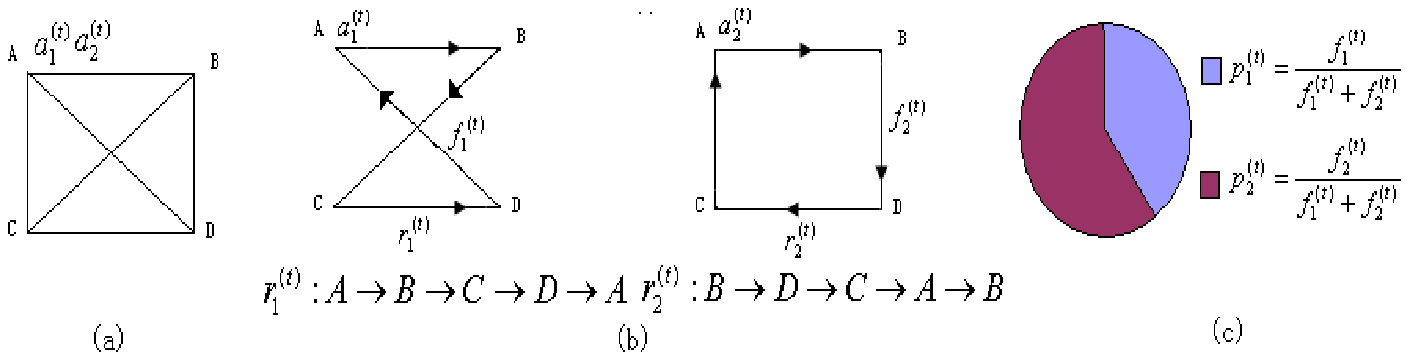,width=10cm,}
\caption{\textbf{The Schematic of pheromone probability }$p_{i}^{(t)}$%
\textbf{. }(a) There is a complete graph with four vertices, two ants $%
a_{1}^{(t)}$ and $a_{2}^{(t)}$ act on it at the $t-th$ iteration
step. (b) Two ants $a_{1}^{(t)}$ and $a_{2}^{(t)}$ select two routes
$r_{1}^{(t)}$ and $r_{2}^{(t)}$ respectively. Every edge of route
$r_{1}^{(t)}$ contains pheromone , $f_{1}^{(t)}$ represents the sum
of pheromone on all edges of route $r_{1}^{(t)}$. As same,
$f_{2}^{(t)}$ represents the sum of pheromone on all edges of route
$r_{2}^{(t)}$. (c) There are two routes $r_{1}^{(t)}$ and
$r_{2}^{(t)}$ totally, $p_{i}^{(t)}$ represents the ratio of
pheromone assigned at route $r_{i}^{(t)}$. The more bigger ratio
$p_{i}^{(t)}$ is, the more possibly the edges at route $r_{i}^{(t)}$
were selected by ants at the next iteration step. That is,
$p_{i}^{(t)}$ is a probability which will affect the route selection
of ants at the next iteration.} \label{figPheromoneProbability_1}
\end{figure}

In formula (\ref{eq6}), $\overset{m}{\underset{j=1}{\sum }}f_{j}^{(t)}$\
represents the sum of pheromone of all routes. $p_{i}^{(t)}$\ represents the
ratio of pheromone that is assigned at the $i-th$ route $r_{i}^{(t)}$. The
more big the ratio $p_{i}^{(t)}$ is, the more possibly the edges of route $%
r_{i}^{(t)}$ are selected by ants at the next iteration step. That is, $%
p_{i}^{(t)}$ is a probability which will affect the route selection of ant
at the next iteration step. $p_{i}^{(t)}$\ is called as \emph{pheromone
probability}, and Fig.\ref{figPheromoneProbability_1} diagrammatizes it.

\bigskip \textbf{Definition 2 (Route Length Set):} At the $t-th$ iteration
step, $m$ ants select $m$ routes. The set of route lengths is denoted as

\begin{equation*}
L\_Set^{(t)}=\{L_{1}^{(t)},L_{2}^{(t)},...,L_{i}^{(t)},...,L_{m}^{(t)}\}
\end{equation*}

\textbf{Definition 3 (Pheromone Probability Set):} The set of pheromone
probabilities is defined as

\begin{equation*}
P\_Set^{(t)}=\{p_{1}^{(t)},p_{2}^{(t)},...,p_{i}^{(t)},...,p_{m}^{(t)}\}
\end{equation*}

\subsection{Statistical Features of Route Length Set}

At the $t-th$ iteration step, the $i-th$ ant $a_{i}$ selects route $%
r_{i}^{(t)}$, where $i=1,2,...,m$. And route $r_{i}^{(t)}$ has length $%
L_{i}^{(t)}$. It's possible that two different ants select same route and
they have same route length. Even it's possible that, two different ants
select two\ different routes, but their lengths are same. Thus, for a given
value of route length $x$, there are set $A_{t}(x)=\{j|L_{j}^{(t)}=x\}$,
where $j$ is the subscript of route $r_{j}^{(t)}$. And set $A_{t}(x)$ is the
set of subscripts of routes which lengths are equal to a given value $x$.
Let $|A_{t}(x)|$ denote the number of elements of set $A_{t}(x)$. Number $%
|A_{t}(x)|$ represents the frequency that the routes with length $x$ being
selected by ants. And real number $\frac{|A_{t}(x)|}{m}$ is the
approximation of probability that represents the degree of possibility that
the routes with length $x$ being selected by ants.

Let
\begin{eqnarray*}
h_{t}(x) &=&\frac{|A_{t}(x)|}{m} \\
x &\in &[L_{inf},L_{sup}]
\end{eqnarray*}%
, where $L_{inf}=min\{L_{i}^{(t)}\}$ and $L_{sup}=max\{L_{i}^{(t)}\}$.

Then $h_{t}(x)$ is the function of probability in theory, and the domain of
function is extended to the set of possitive real numbers in general.

To observer the statistical feature of route length set $L\_Set^{(t)}$, its
histogram is plotted, which is the approximation of probability function $%
h_{t}(x)$. The method of plot is proposed as below:

Firstly, a two-dimensional coordinate frame is constructed, the $x-axis$
denotes the value of route length $x$, and the $y-axis$ denotes probability $%
h_{t}(x)$. The $x-axis$ is divided into equal intervals, and the size of
each interval is denoted by $\delta $.

Secondly, calculate the approximation of probability $h_{t}(x)$ for every
interval $I$ : Suppose interval $I$ has a counter $c$ which initial value is
set to zero (i.e., $c=0$). When a length $L_{i}^{(t)}$ falling into this
interval (i.e., $L_{i}^{(t)}\in I$), let the counter add one (i.e., $%
c\leftarrow c+1$). Then for arbitrary $x\in I$, there is function value $%
\frac{c}{m}$. Function value $\frac{c}{m}$ is the approximation of
probability $h_{t}(x)$. Under the condition that the size of interval $I$
becomes very small, we have $h_{t}(x)=\frac{c}{m}$.

\begin{figure}[tbh]
\epsfig{file=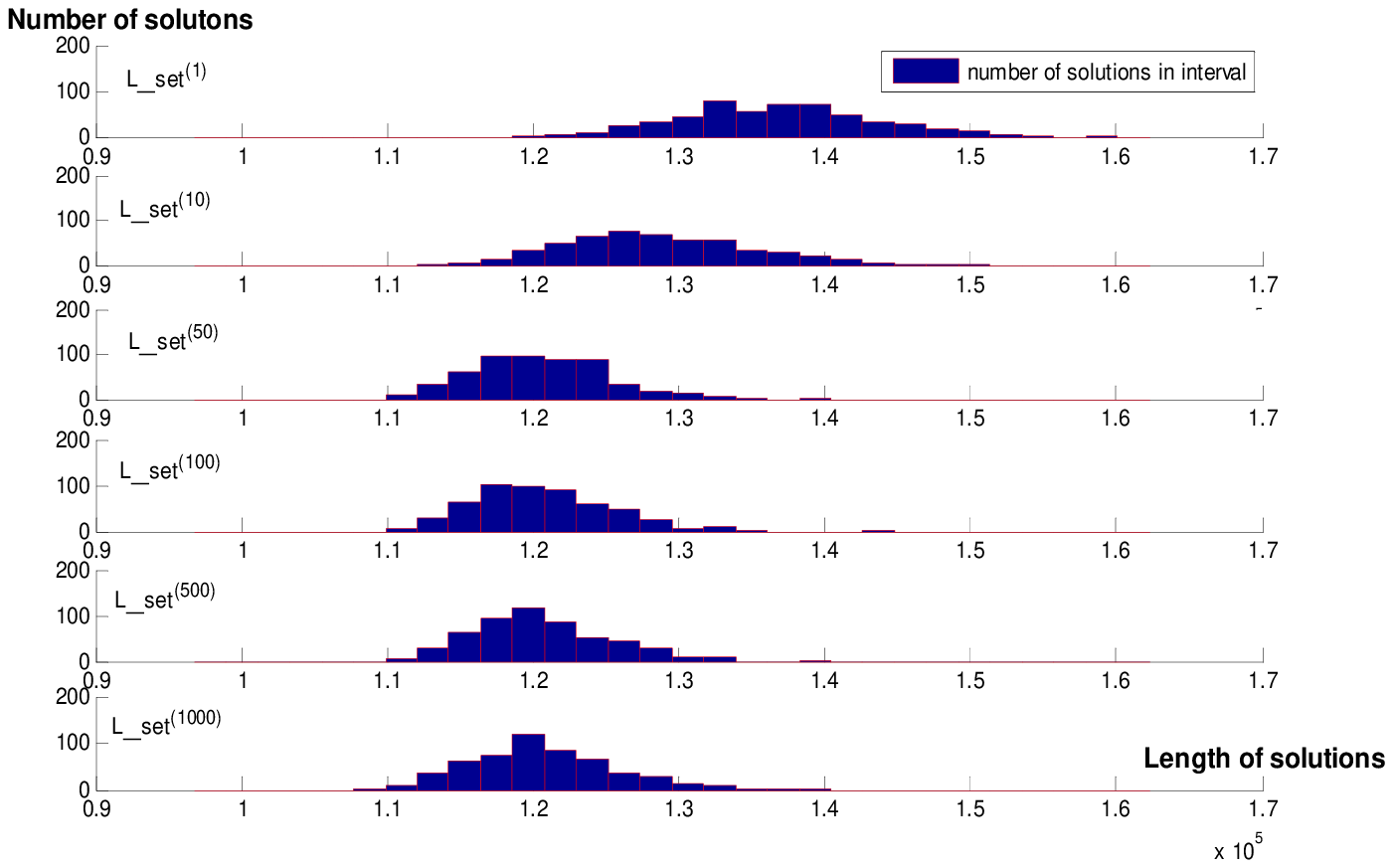,width=10cm,}
\caption{\textbf{The Histogram of the Route Length Set }$L\_Set^{(t)}$%
\textbf{: }$X-axis$ represents the value of route length $x$, $Y-axis$
represents the probability that the routes with length $x$ being selected by
ants . In this figure, the probability is replaced by frequency for
direct-viewing. This figure shows that set $L\_Set^{(t)}$ has statistical
feature. The distribution of set $L\_Set^{(t)}$ is convergent. Notice: The
test data is pr136, \ number of cities and ants is 136 and 544 respectively.
The $x-axis$ is divided into equal interval with size 2183 (i.e.$\protect%
\delta =2183$ ). The histogram of at $t-th$ iteration is shown here, where $%
t=1,10,50,100,500,1000$. The same features are also found in other test
data, such as pr107, d198, pr226, d493 and so on. All test data in this
paper is downloaded from
http://www.iwr.uniheidelberg.de/iwr/comopt/soft/TSPLIB95/TSPLIB.html }
\label{figHistogram}
\end{figure}

The histogram of test data pr136 is shown at Fig.\ref{figHistogram}. Fig.\ref%
{figHistogram} demonstrates that route length set $L\_Set^{(t)}$ has some
statistical features, and they are summarized as below.

(1) The value of route length $x$ is random data and has probability $%
h_{t}(x)$. The expectation and deviation of set $L\_Set^{(t)}$ exist.

(2) Being big in the middle and small at both sides, that is the shape of
probability function $h_{t}(x)$. It's a typical distribution feature.

(3) With the increase of iteration step (i.e., $t\rightarrow t+1$ ), the
distribution of set $L\_Set^{(t)}$ will become stable. That is, the sequence
of probability functions $\{h_{1}(x),h_{2}(x),...,h_{t}(x),...\}$ is
convergent.

\subsection{The Expectation and Deviation of Route Length Set}

Expectation and deviation are the two most essential characteristics of
distribution, these two characteristics of set $L\_Set^{(t)}$ will be
calculated in this section. The expectation and standard deviation of set $%
L\_Set^{(t)}$ are denoted by $\overset{-}{L}^{(t)}$and $\sigma ^{(t)}$
respectively in this paper. $\overset{-}{L}$ and $\sigma ^{(t)}$ are defined
as below:

\textbf{Definition 4} (the expectation of set $L\_Set^{(t)}$):

\begin{equation}
\overset{-}{L}^{(t)}=\frac{1}{m}\overset{m}{\underset{i=1}{\sum L_{i}^{(t)}}}
\label{eq7}
\end{equation}

, where $m$ is the number of ants.

\textbf{Definition 5} (the standard deviation of set $L\_Set^{(t)}$):

\begin{equation}
\sigma ^{(t)}=\sqrt{\frac{1}{m}\overset{m}{\underset{i=1}{\sum }}%
|L_{i}^{(t)}-\overset{-}{L}^{(t)}|^{2}}  \label{eq8}
\end{equation}

\begin{figure}[tbh]
\epsfig{file=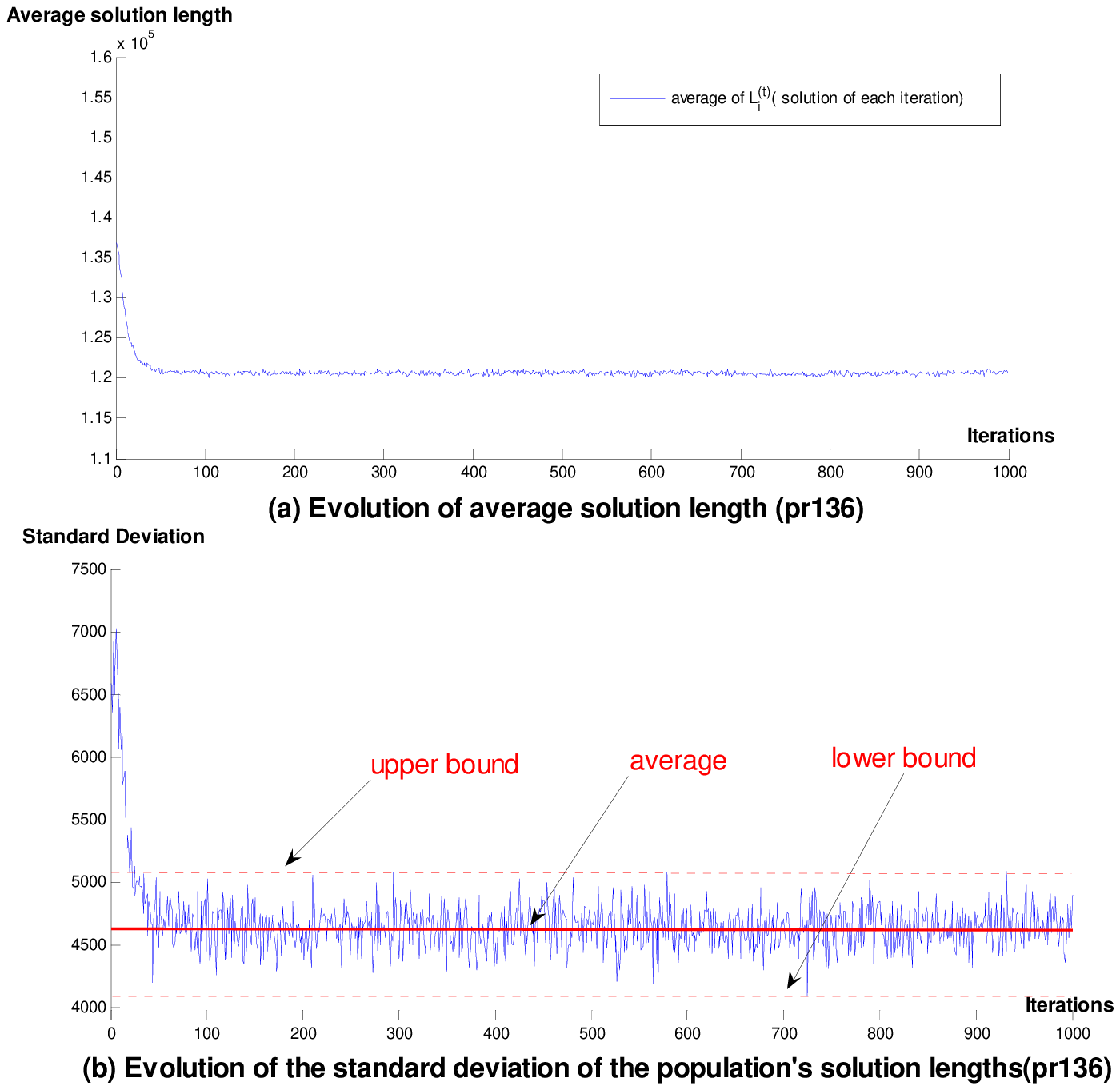,width=10cm,}
\caption{\textbf{The Feature of Expectation and Standard Deviation of Route
Length Set:} Figure (a) and (b) diagrammatize the two curves of sequence $\{%
\protect\overset{-}{L}^{(1)},\protect\overset{-}{L}^{(2)},...,%
\protect\overset{-}{L}^{(t)},...\}$ and $\{\protect\sigma ^{(1)},\protect%
\sigma ^{(2)},...,\protect\sigma ^{(t)},...\}$\ respectively, where $%
\protect\overset{-}{L}^{(t)}$and $\protect\sigma ^{(t)}$\ are the
expectation and standard deviation of route length set $L\_Set^{(t)}$\
respectively. Figure (a) shows that expectation descends continuously and
converges to a constant value. Figure (b) shows that all standard deviations
fluctuate narrowly in a small interval comparing with average route length
and most of deviations are close to a constant value. The similar feature
are observed from other instances, such as pr107, d198, pr226, d493 and so
on. Notice: The test data in this figure is pr136, the number of cities is $%
n=136$, the number of ants is $m=544$, and the number of maximum iteration
is $N_{C_{\max }}=1000$.}
\label{figAveDev}
\end{figure}

Two sequences $\{\overset{-}{L}^{(1)},\overset{-}{L}^{(2)},...,\overset{-}{L}%
^{(t)},...\}$ and $\{\sigma ^{(1)},\sigma ^{(2)},...,\sigma ^{(t)},...\}$
are shown at Fig.\ref{figAveDev}. The subfigure (a) of Fig.\ref{figAveDev}
shows that expectation $\overset{-}{L}^{(t)}$descends continuously and
converges to a constant value. The subfigure (b) shows that all standard
deviations fluctuate narrowly in an interval and most of deviations are
close to a constant value.

\subsection{Holding View Point of Statistics to Understand ACO Convergence}

There are three types of understanding for the convergence of ACO:

\textbf{Type 1:} With the increase of iteration steps, all ants will select
the optimal route which has shortest length.

\textbf{Type 2:} With the increase of iteration steps, all ants will select
a unique fixed (or stable) route, but it is not optimal possibly.

\textbf{Type 3:} With the increase of iteration steps, more than one fixed
routes are selected by different ants. That is, ACO converges to a stable
set which consists of some fixed routes, not a unique route.

\bigskip

ACO converging to optimal route is difficult in general, the first type is
not common in practice. The 2nd type is also not common in practice, and it
never be observed in the authors' experiment. Instead of the 2nd type, the
3rd type is common in practice. For example, Fig.\ref{figHistogram} shows
that, there are always different routes selected by ants at every iteration
step, and the convergent route is not unique. \textbf{Since the 3rd type is
common and has more practical worthiness, the convergence of ACO refers to
this type in this paper.}

In addition, the aim of ACO is to find the shortest route length, and the
difference of the routes is not cared. Therefore, a equivalent statement of
the 3rd type is that, ACO converges to a stable set which consists of stable
route lengths .

According to above discusion, if ACO converges, the stable set will appear,
which consists of stable route lengths. Then the histogram of this stable
set is convergent (see Fig. \ref{figHistogram}). That is, ACO converging
results in probability sequence $\{h_{1}(x),h_{2}(x),...,h_{t}(x),...\}$
being convergent. At the same time, sequence $%
\{h_{1}(x),h_{2}(x),...,h_{t}(x),...\}$ being convergent will result in ACO
converging also, and it is proved as below:

Let set $A_{t}(x)=\{j|L_{j}^{(t)}=x\}$. Then $h_{t}(x)=\frac{|A_{t}(x)|}{m}$%
. Thus, if $\{h_{1}(x),h_{2}(x),...,h_{t}(x),...\}$ is convergent, $%
|A_{t}(x)|$ becomes fixed (or stable). Since route $r_{j}^{(t)}$ represents
the route selected by ant $a_{j}$, number $|A_{t}(x)|$ represents the number
of ants which routes has length $x$. There are only two factors to cause $%
|A_{t}(x)|$ becoming fixed. One factor is ACO being convergent. The other
factor is that, some ants coming into set $A_{t}(x)$ and some coming out,
the quantities of input and output are equal. The second factor is too
special so that it does not exist. Therefore, if $%
\{h_{1}(x),h_{2}(x),...,h_{t}(x),...\}$ is convergent, ACO will be
convergent.

According to above discussion, the following conclusion is obtained:

\textbf{Conclusion 1:} ACO being convergent is equivalent to the sequence of
probability functions $\{h_{1}(x),h_{2}(x),...,h_{t}(x),...\}$ being
convergent.

This conclusion shows that, the histogram of route length set becoming
convergent is the marker of ACO being convergent (see Fig.\ref{figHistogram}%
).

\section{Using Pheromone Probability to Observe The Statistical Feature of
Route Length Set}

\subsection{The Pseudo-Probability and Pseudo-Histogram of Route Length Set}

\textbf{The definition of pseudo-probability }$h_{t}^{^{\prime }}(x)$\textbf{%
:}

At the $t-th$ iteration step, every ant will select a route. The $i-th$ ant $%
a_{i}$ selects route $r_{i}^{(t)}$, and $r_{i}^{(t)}$ has length $%
L_{i}^{(t)} $, where $i=1,2,...,m$. Each route $r_{i}^{(t)}$ contains the
amount of pheromone $f_{i}^{(t)}$, which is the sum of pheromone depositing
on every edge of route $r_{i}^{(t)}$. The pheromone probability is the ratio
of pheromone, it is defined as $p_{i}^{(t)}=\frac{\ f_{i}^{(t)}}{\overset{m}{%
\underset{j=1}{\sum }}f_{j}^{(t)}}$.

Set $A_{t}(x)=\{j|L_{j}^{(t)}=x\}$ is the subscripts set of routes which
length is a given value $x$. Basing on set $A_{t}(x)$, \textbf{%
pseudo-probability} is defined as

\begin{eqnarray*}
h_{t}^{^{\prime }}(x) &=&\underset{j\in A_{t}(x)}{\sum p_{j}^{(t)}} \\
x &\in &[L_{inf},L_{sup}]
\end{eqnarray*}%
, where $L_{inf}=min\{L_{i}^{(t)}\}$ and $L_{sup}=max\{L_{i}^{(t)}\}$%
.\bigskip

Pseudo-probability $h_{t}^{^{\prime }}(x)$ is the sum of pheromone
probabilities which associated route has length $x$.

\bigskip

\textbf{The pseudo-histogram of route length set }$L\_Set^{(t)}$\textbf{:}

Pseudo-histogram is also a histogram in which pseudo-probability $%
h_{t}^{^{\prime }}(x)$ replace probability $h_{t}(x)$ to estimate the
distribution of route length set $L\_Set^{(t)}$. It is generated by
following method:

Firstly, a two-dimensional coordinate frame is constructed, the $x-axis$
denotes the value of route length $x$, and the $y-axis$ denotes
pseudo-probability $h_{t}^{^{\prime }}(x)$. The $x-axis$ is divided into
equal intervals, and the size of each interval is denoted by $\delta $.

Secondly, calculate the approximation of pseudo-probability for every
interval $I$: Suppose $x$ represents argument and $x\in I$. A counter $d$ is
attached to interval $I$, and its initial value is set to zero (i.e., $d=0$%
;).\ If $L_{i}^{(t)}$\ falls into interval $I$ (i.e., $L_{i}^{(t)}\in I$),
its associated pheromone probability $p_{i}^{(t)}$ is add to $d$ (i.e., $%
d=d+p_{i}^{(t)}$). The value $d$ is the function value of argument $x$. When
the size of interval $I$ limits to zero ideally, the value $d$ limits to
pseudo-probability $h_{t}^{^{\prime }}(x)$.

\bigskip

\begin{figure}[tbh]
\caption{\textbf{The Pseudo-Histogram of Route Length Set }$L\_Set^{(t)}$%
\textbf{\ Calculated by Pheromone Probability }$p_{i}^{(t)}$\textbf{.} The $%
X-axis$ denotes the value of route length $x$ and is divided into small
intervals. The value at $Y-axis$ denotes the pseudo-probability $%
h_{t}^{^{\prime }}(x)$, which is the sum of all pheromone probabilities $%
p_{i}^{(t)}$ under the condition $L_{i}^{(t)}=x$. The input data of this
instance is pr136, where the number of cities $n=136$, number of ants $m=544$%
\ and length of interval $\protect\delta =2183$. Comparing with Figure.%
\protect\ref{figHistogram}, this figure is similar to it. And this
similarity excites the guess pseudo-probability $h_{t}^{^{\prime }}(x)$ is
the approximation of probability $h_{t}(x)$, which represents the degree of
possibility that the routes with length $x$ being selected by ants.}
\label{figRelation}\epsfig{file=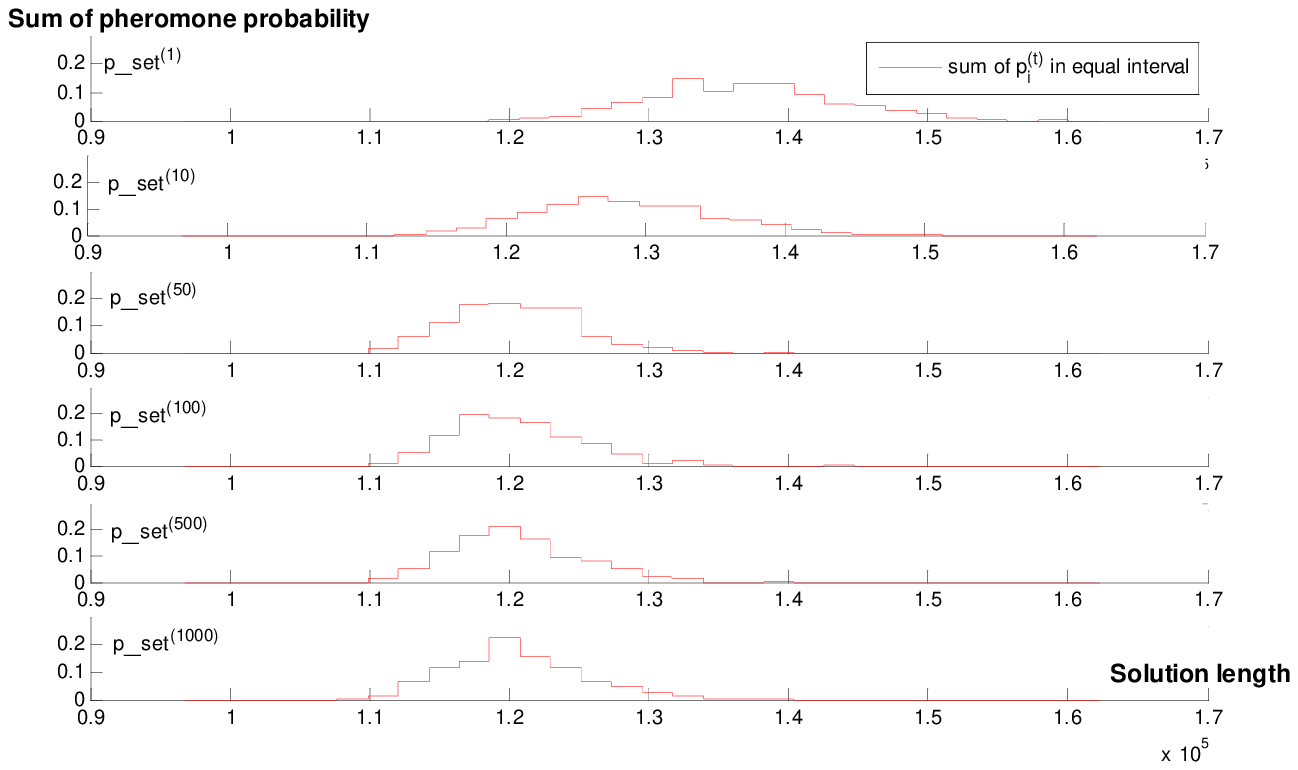,width=12cm,}
\end{figure}

The pseudo-histogram is shown in Fig.\ref{figRelation}. Comparing with the
histogram of probability function shown at Fig.\ref{figHistogram},
pseudo-histogram is very similar to it. Probability $h_{t}(x)$ represents
the degree of possibility that the routes with length $x$ being selected by
ants, pseudo-probability $h_{t}^{^{\prime }}(x)$ represents the sum of
pheromone probability $p_{i}^{(t)}$ which associated route $r_{i}^{(t)}$ has
length $x$. The similarity of these two figures excites the guess that
pseudo-probability is the approximation of probability(i.e., $%
h_{t}(x)\thickapprox h_{t}^{^{\prime }}(x)$).

\subsection{Pseudo-Expectation $\protect\overset{-}{L}^{^{\prime }(t)}$and
Pseudo-Deviation $\protect\sigma ^{^{\prime }(t)}$}

\textbf{Definition 6} ($\overset{-}{L}^{^{\prime }(t)}$, Pseudo-Expectation
of Route Length Set $L\_Set^{(t)}$ Calculated by Pheromone Probability):

\begin{equation}
\overset{-}{L}^{^{\prime }(t)}=\underset{x\in V}{\sum }xh_{t}^{^{\prime
}}(x)=\overset{m}{\underset{i=1}{\sum }}L_{i}^{(t)}\times p_{i}^{(t)}
\label{eq9}
\end{equation}%
, where $x$ denotes the value of route length and $V$ denotes the set of
these values.

\textbf{Definition 7} ($\sigma ^{^{\prime }(t)}$, Pseudo-Deviation
Calculated by Pheromone Probability):

\begin{equation}
\sigma ^{^{\prime }(t)}=\sqrt{\overset{m}{\underset{i=1}{\sum }p_{i}^{(t)}}%
|L_{i}^{(t)}-\overset{-}{L}^{^{\prime }(t)}|^{2}}  \label{eq10}
\end{equation}

\begin{figure}[tbh]
\epsfig{file=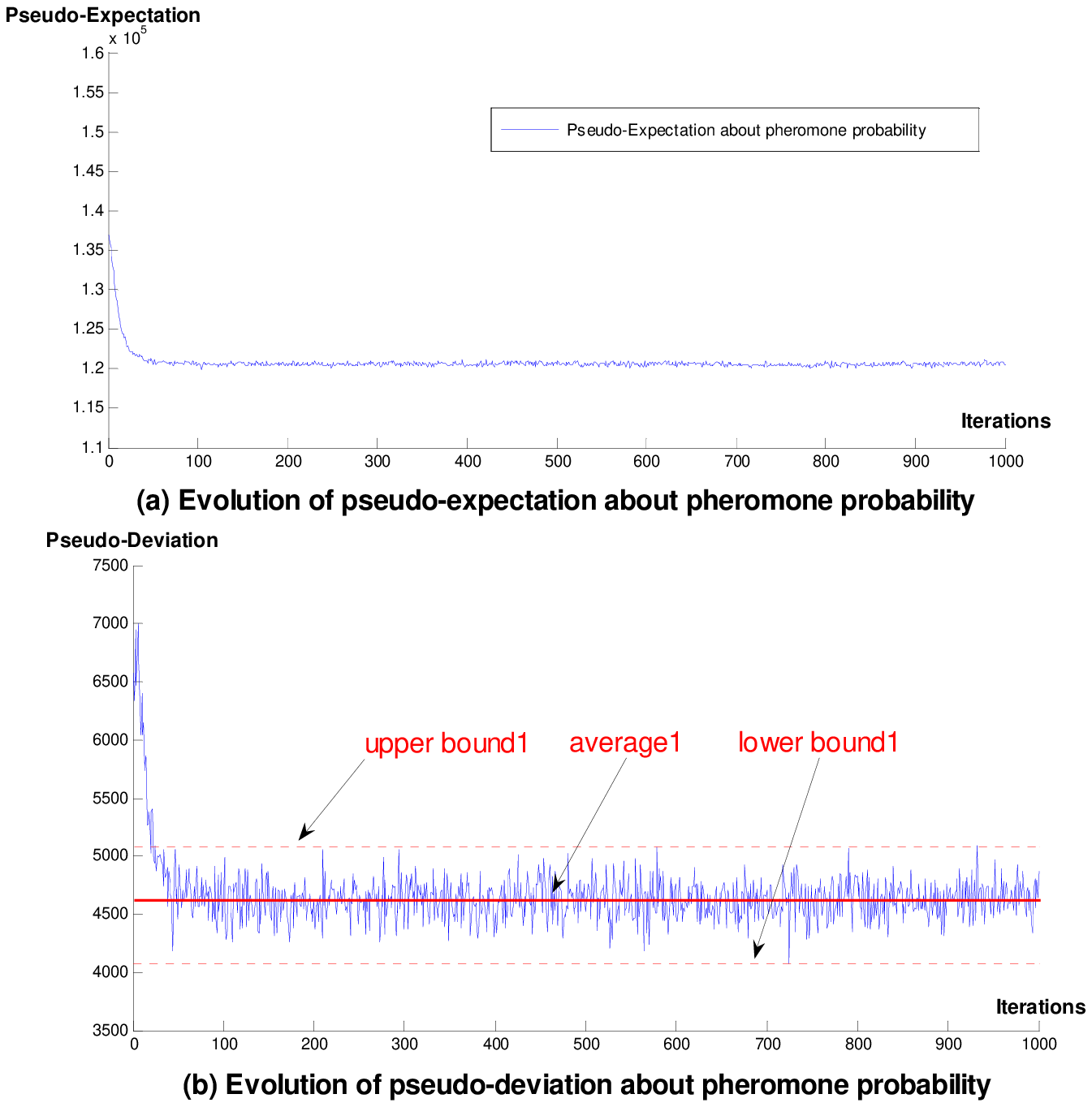,width=10cm,}
\caption{\textbf{Pseudo-Expectation }$\protect\overset{-}{L}^{^{\prime }(t)}$%
\textbf{and Pseudo-Deviation }$\protect\sigma ^{^{\prime }(t)}$\textbf{\ :}
Comparing with Fig.\protect\ref{figAveDev}, this figure is similar to it.
\textit{This similarity provides a evidence to support conclusion 2: }$%
h_{t}(x)\thickapprox h_{t}^{^{\prime }}(x)$.\textit{\ }Notice: The input
data of this shown instance is pr136, the number of ants $m=544$, the number
of cities $n=136$ and the number of maximum iteration $N_{C_{\max }}=1000$.
The same conclusion is also found in other test data, such as pr107, d198,
pr226, d493 and so on.}
\label{figExVar}
\end{figure}

\begin{figure}[tbh]
\epsfig{file=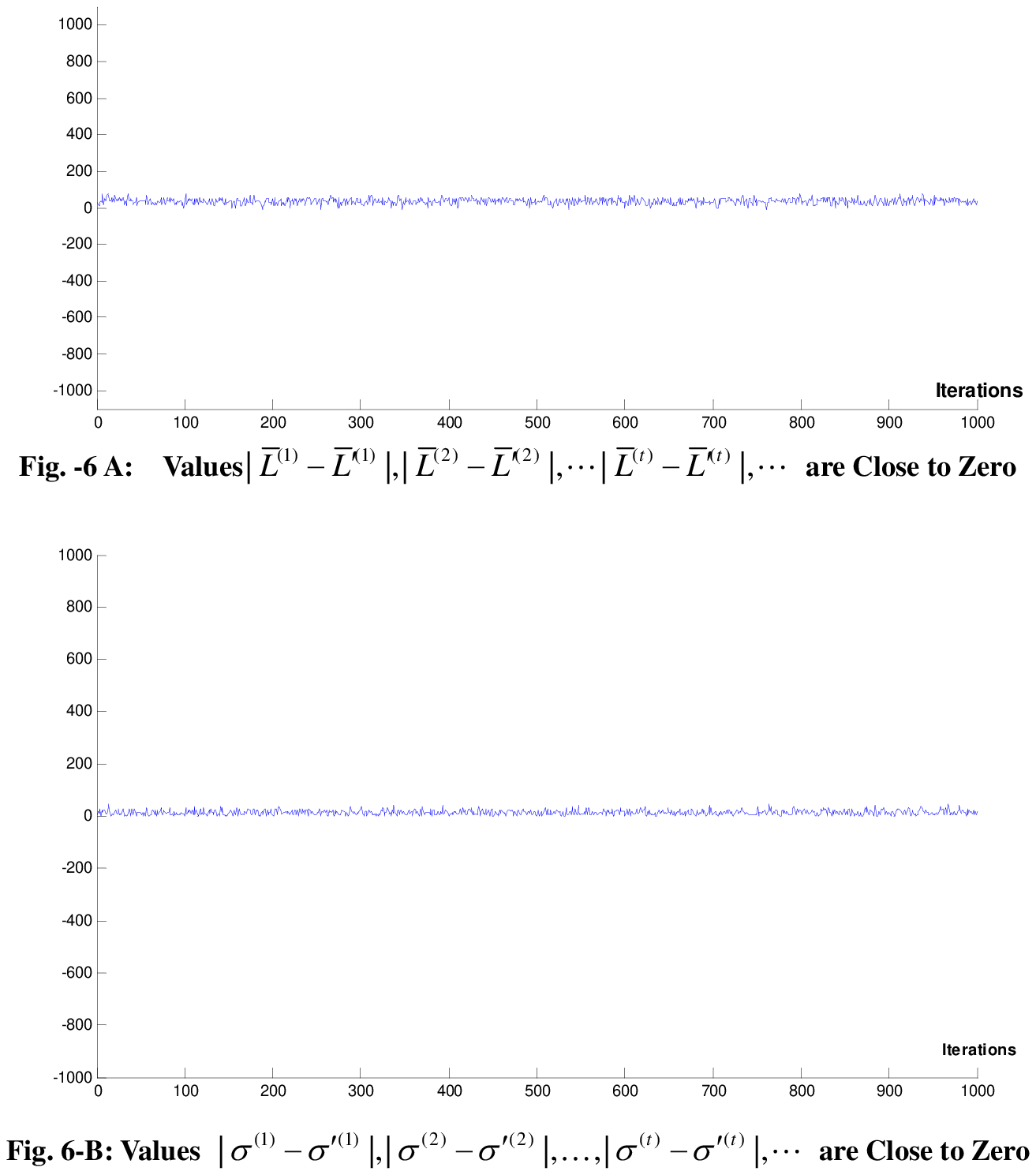,width=10cm}
\caption{\textbf{Sequence of Pesudo-Expectation }$\{\protect\overset{-}{L}%
^{^{\prime }(t)}\}$\textbf{\ and Pesudo-Deviation }$\{\protect\overset{-}{%
\protect\sigma }^{^{\prime }(t)}\}$\textbf{\ Is Close to }$\protect\overset{-%
}{\{L}^{(t)}\}$\textbf{\ and }$\{\protect\overset{-}{\protect\sigma }%
^{(t)}\} $\textbf{\ Respectively.} Two sequences \{$|\protect\overset{-}{L}%
^{(t)}-\protect\overset{-}{L}^{^{\prime }(t)}|$\} and \{$|\protect\sigma %
^{(t)}-\protect\sigma ^{^{\prime }(t)}|$\} are shown in this figure. This
figure shows that $\protect\overset{-}{L}^{(t)}\thickapprox \protect\overset{%
-}{L}^{^{\prime }(t)}$and $\protect\sigma ^{(t)}\thickapprox \protect\sigma %
^{^{\prime }(t)}$. This evidence further supports conclusion 2.
Notice: The experiment parameters are same to Fig.5}
\label{figSubstraction-1}
\end{figure}

The two sequence \{$\overset{-}{L}^{^{\prime }(1)},\overset{-}{L}^{^{\prime
}(2)},...,\overset{-}{L}^{^{\prime }(t)},...$\} and sequence \{$\sigma
^{^{\prime }(1)},\sigma ^{^{\prime }(2)},..,\sigma ^{^{\prime }(t)},...$\}
are shown at Fig.\ref{figExVar}. Comparing with Fig.\ref{figAveDev}, Fig.\ref%
{figExVar} is very similar to it. Two sequences \{$|\overset{-}{L}^{(t)}-%
\overset{-}{L}^{^{\prime }(t)}|$\} and \{$|\sigma ^{(t)}-\sigma ^{^{\prime
}(t)}|$\} are shown in Fig.\ref{figSubstraction-1}, this figure demonstrates
that $\overset{-}{L}^{(t)}\thickapprox \overset{-}{L}^{^{\prime }(t)}$and $%
\sigma ^{(t)}\thickapprox \sigma ^{^{\prime }(t)}$. Expectation and
deviation are two most important characteristics of set of random data. And
Fig.\ref{figExVar} and Fig.\ref{figSubstraction-1} are two evidences to
support the conclusion
\begin{equation*}
h_{t}(x)\thickapprox h_{t}^{^{\prime }}(x)
\end{equation*}%
, where $h_{t}(x)$ and $h_{t}^{^{\prime }}(x)$ denotes the probability and
pseudo-probability respectively.

Since histogram and pseudo-histogram is very similar and $\overset{-}{L}%
^{(t)}\thickapprox \overset{-}{L}^{^{\prime }(t)}$and $\sigma
^{(t)}\thickapprox \sigma ^{^{\prime }(t)}$, we have following conclusion:

\textbf{Conclusion 2:} With the increasing of iteration step,
pseudo-probability is the approximation of probability (i.e., $%
h_{t}(x)\thickapprox h_{t}^{^{\prime }}(x)$ when $t\rightarrow \infty $).

\bigskip

Conclusion 1 shows probability function $h_{t}(x)$ being convergent is
equivalent to ACO being convergent. Since $h_{t}(x)\thickapprox
h_{t}^{^{\prime }}(x)$, we have

\bigskip

\textbf{Conclusion 3:} Pseudo-probability $h_{t}^{^{\prime }}(x)$ being
convergent is equivalent to ACO being convergent

\bigskip

Pseudo-probability $h_{t}^{^{\prime }}(x)$ being convergent results in ACO
being convergent. And when ACO being convergent, every route selected by ant
is fixed. This situation results in the amount of pheromone depositing on
convergent route is fixed and its ratio (i.e., pheromone probability $%
p_{i}^{(t)}$) is fixed too. Therefore, function $h_{t}^{^{\prime }}(x)$
being convergent results in pheromone probability $p_{i}^{(t)}$ being
convergent, where $i=1,2,...,m$. On the other hand, pheromone probability $%
p_{i}^{(t)}$ being convergent results in the function $h_{t}^{^{\prime }}(x)$
being convergent and ACO being convergent. Then, we have

\bigskip

\textbf{Conclusion 4: }Pseudo-probability $h_{t}^{^{\prime }}(x)$\ being
convergent is equivalent to pheromone probability set $P\_Set^{(t)}$\ being
convergent, where the convergence of $P\_Set^{(t)}$\ refers to that every
pheromone probability in this set is convergent.

\textbf{Conclusion 5: }Pheromone probability set $P\_Set^{(t)}$\ being
convergent is equivalent to ACO being convergent.

\section{Entropy Convergence}

\subsection{Entropy of Pheromone and Its Convergence}

In 1948 Shannon introduced the entropy \cite{Shannon} into information
theory for the first time. In information theory, entropy is a measure of
the uncertainty associated with random system. The lower entropy is, the
lower the uncertainty of system is. Entropy is defined as

\begin{equation}
H=-\overset{n}{\underset{i=1}{\sum }}p_{i}\cdot log_{_{2}}p_{i}  \label{eq11}
\end{equation}%
, where $p_{i}$ denotes the probability.

At $t-th$ iteration of ACO, ant $a_{i}$ select route $r_{i}^{(t)}$, where $%
i=1,2,...,m$. Route $r_{i}^{(t)}$ associates with pheromone probability $%
p_{i}^{(t)}$, which is the ratio of pheromone assigned at route $r_{i}^{(t)}$%
. All pheromone probability $p_{i}^{(t)}$ comprise set $P\_Set^{(t)}=%
\{p_{1}^{(t)},p_{2}^{(t)},...,p_{i}^{(t)},...,p_{m}^{(t)}\}$.

According to Eq.\ref{eq11}, \textbf{entropy of pheromone} is defined as

\begin{equation}
H(P\_Set^{(t)})=-\overset{n}{\underset{i=1}{\sum }}p_{i}^{(t)}\cdot
log_{_{2}}p_{i}^{(t)}  \label{eq12}
\end{equation}

It is simplified as

\begin{equation}
H_{t}=-\overset{n}{\underset{i=1}{\sum }}p_{i}^{(t)}\cdot
log_{_{2}}p_{i}^{(t)}  \label{eq13}
\end{equation}

Pheromone probability $p_{i}^{(t)}$ represents the ratio of pheromone
assigned at route $r_{i}^{(t)}$. If every route is assigned equal amount of
pheromone, all ants don't know which route is best and select route
randomly. At this time, all pheromone probabilities are equal (i.e., $%
p_{i}^{(t)}=\frac{1}{m}$), entropy of pheromone is maximum, and the
uncertainty degree that ants selecting route is maximum. And this situation
is often happened at the early iteration steps of ACO. If pheromone is
assigned at few routes, most of ants will select these routes with high
probability. At this time, there is low uncertainty for ants selecting
route, entropy is small. This situation is often happened at the iteration
steps at which ACO is close to convergence.

\begin{figure}[tbh]
\epsfig{file=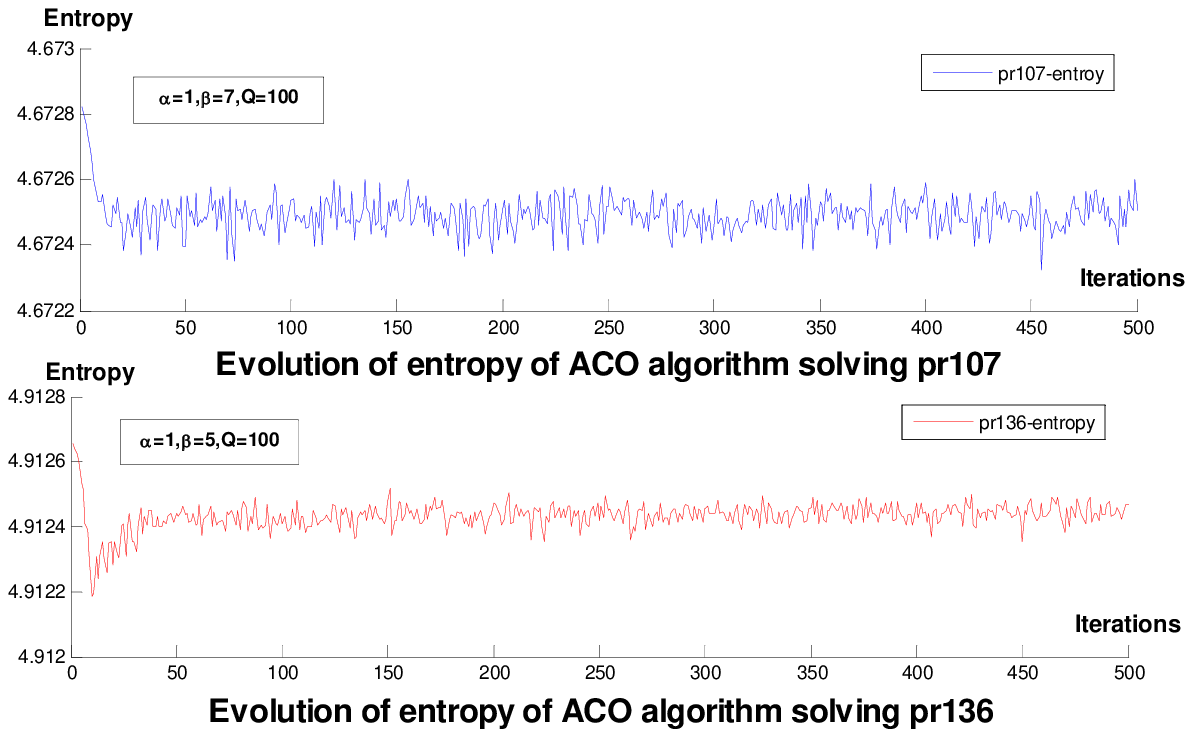,width=10cm,}
\caption{\textbf{Entropy of Pheromone Is Convergent.} Two curves show that
two entropy sequences are convergent approximately. And the amplitude of
swing is very narrow, it is less than $0.0002/4.6$ or $0.0004/4.9$ . Notice:
The shown test data are pr107 and pr136, the number of ant is equal to the
number of cites (i.e. $m=107$ and $m=136$), the number of maximum iteration
is $N_{C_{\max }}=500$ and the feature of entropy convergence is observed
from other test data also, such as d198, pr226, d493 and so on. The entropy
is calculated by $H(t)=-\protect\overset{n}{\protect\underset{i=1}{\sum }}%
p_{i}^{(t)}\cdot lnp_{i}^{(t)}$ in this figure.}
\label{figEntropy}
\end{figure}

With the increase of iteration step, every route selected by ants will
become fixed, the pheromone depositing on it becomes fixed (stable) and its
pheromone probability becoming fixed too. This situation results in the
sequence $\{H_{1},H_{2},...,H_{t},...\}$ converging. The test result at Fig.%
\ref{figEntropy} shows that the entropy sequence is convergent.

\subsection{Entropy Convergence Is A Marker of ACO Convergence}

Entropy is the most essential characteristics of a random system. Thus, the
convergence of entropy sequence $\{H_{1},H_{2},...,H_{t},...\}$\ is the
marker of the convergence of set $P\_Set^{(t)}$. Set $P\_Set^{(t)}$ being
convergent is equivalent to ACO being convergent according to conclusion 5.
Therefore, the convergence of entropy sequence $\{H_{1},H_{2},...,H_{t},...\}
$ is the marker of the convergence of ACO. When ACO is convergent, set $%
P\_Set^{(t)}$ is convergent, entropy sequence is convergent too. If ACO is
not convergent, set $P\_Set^{(t)}$ is not convergent, entropy sequence is
not convergent too. On the other hand, when entropy sequence is convergent,
set $P\_Set^{(t)}$ is convergent very possibly because entropy is its
essential characteristic, ACO is convergent too. If entropy sequence is not
convergent, set $P\_Set^{(t)}$ is not convergent very possibly, ACO is not
convergent too.

\emph{Therefore, the convergence of entropy sequence is a marker of minimum
iteration steps at which ACO is convergent.}

In addition, the convergence of entropy sequence $%
\{H_{1},H_{2},...,H_{t},...\}$ has usual \ criterion $\frac{|H_{t}-H_{t-1}|}{%
H_{t-1}}<\varepsilon $ \cite{Pang}. And criterion$\frac{|H_{t}-H_{t-1}|}{%
H_{t-1}}<\varepsilon $ is a very simple criterion to estimate the minimum
iteration number at which ACO is convergent possibly.

\section{Application of Entropy Convergence}

\subsection{Apply Entropy Convergence as Termination Criterion of ACO}

The improved ACO algorithm with criterion $\frac{|H_{t}-H_{t-1}|}{H_{t-1}}%
<\varepsilon $ is presented as below, and it is named \textbf{ACO-Entropy}
in this paper.

\textbf{Algorithm ACO-Entropy}

\textbf{Step1.} Initialize pheromone trails for all edges and put $m$ ants
at different cities. Let $t=0$ , $\Delta \tau _{ij}(0)=0$ and $H_{0}=\log
_{2}m$.

\textbf{Step2. do}

\{

\ \ \ \textbf{Step2.1} $t\leftarrow t+1$.

\ \ \ \textbf{Step2.2} The ants choose next cities according to transition
probability.

\ \ \ \textbf{Step2.3} After all ants finish their travels, pheromone are
updated.

\ \ \ \textbf{Step2.4} The pheromone probability $p_{i}^{(t)}$ and the
entropy $H_{t}$ are calculated by

\ \ \ \ \ \ \ \ \ \ \ formula \ (\ref{eq6}) and (\ref{eq13}) respectively.

\}while($\frac{|H_{t}-H_{t-1}|}{H_{t-1}}\geq \varepsilon $)

\textbf{Step3.} End procedure and output result.

\subsection{The Experiment and Comparison}

All data tested in this paper are downloaded from
http://www.iwr.uniheidelberg.de/iwr/

comopt/soft/TSPLIB95/TSPLIB.html. All algorithms in this paper run on
personal computer, CPU (2): 1.60GHZ, Memory: 480M, Software: Matlab 7.1. All
parameters are set as below:

$\alpha =1$, $\beta =8$, $\rho =0.4$, $Q=100$, $\tau _{ij}(0)=1$, $m=n$, $%
\varepsilon =0.001$, $N_{C_{\max }}=1000$.

To test the performance of ACO-Entropy, two algorithms of ACO and the
ACO-Entropy are tested in this paper, where ACO refers to Ant-Cycle shown at
section II, which is often used standard algorithm.

\begin{figure}[tbh]
\epsfig{file=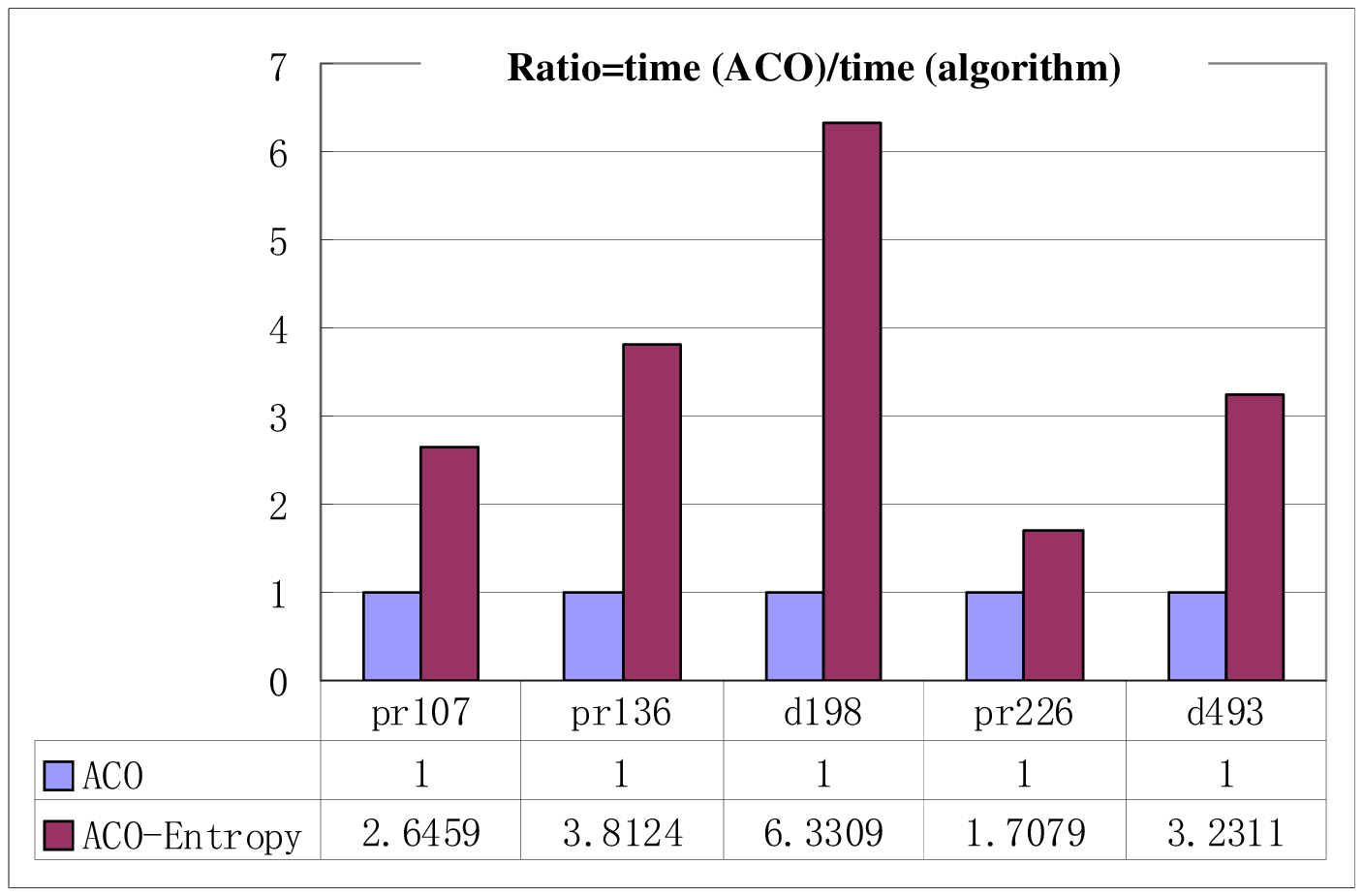,width=10cm,}
\caption{\textbf{Comparison of the Running Speed of ACO and ACO-Entropy}.
ACO-Entropy is faster than ACO by factors of 2-6 under the same condition
and the nearly same quality of solution is obtained (see Table 1), where ACO
refers to Ant-Cycle shown at section II.}
\label{figComparison}
\end{figure}

Table.1 and Fig.\ref{figComparison} show that, ACO-Entropy is faster than
ACO by factors of 2-6 under the same condition and nearly same quality of
solution is obtained.

\begin{center}
{\small \ }%
\begin{tabular}{|c|c|c|c|c|}
\hline
{\small Input} & {\small Number} & \multicolumn{3}{|c|}{\small ACO-Entropy}
\\ \hline
{\small Data} & {\small of Test} & {\small Average Solution} & {\small %
Average Time(s)} & {\small Iteration Number} \\ \hline
{\small pr107} & {\small 10} & {\small 46294} & {\small 163.0804} & {\small %
189} \\ \hline
{\small pr136} & {\small 10} & {\small 108467} & {\small 173.3620} & {\small %
131} \\ \hline
{\small d198} & {\small 10} & {\small 17135} & {\small 447.4071} & {\small %
100} \\ \hline
{\small pr226} & {\small 10} & {\small 84718} & {\small 2466.1} & {\small 293%
} \\ \hline
{\small d493} & {\small 2} & {\small 39851} & {\small 16405} & {\small 155}
\\ \hline
\multicolumn{2}{|c}{} & \multicolumn{3}{|c|}{\small ACO} \\ \hline
{\small pr107} & {\small 10} & {\small 45973} & {\small 431.496} & {\small %
500} \\ \hline
{\small pr136} & {\small 10} & {\small 102608} & {\small 660.918} & {\small %
500} \\ \hline
{\small d198} & {\small 10} & {\small 16891} & {\small 2832.5} & {\small 500}
\\ \hline
{\small pr226} & {\small 10} & {\small 84514} & {\small 4211.8} & {\small 500%
} \\ \hline
{\small d493} & {\small 2} & {\small 38926} & {\small 53007} & {\small 200}
\\ \hline
\multicolumn{5}{|c|}{%
\begin{tabular}{l}
{\small Table1. Performance Comparison of ACO and ACO-Entropy:} \\
{\small This table shows that ACO-Entropy is faster than ACO by factors of
2-6.} \\
{\small The two solution qualities of ACO-Entropy and ACO are nearly same.}%
\end{tabular}%
} \\ \hline
\end{tabular}
\end{center}

\section{Conclusion}

The convergence of ACO is the base of ACO, its study is not much currently.
The convergence under some special conditions has be studied, and the view
point of study are Graph theory, Markov process, and so on. It is
interesting to find a new view point to study ACO convergence under general
condition. The aim of this paper is to explore new view point of studying
ACO convergence under general condition and to find the new marker of ACO
convergence.

Since ACO is kind of probabilistic algorithm, the feature of its convergence
possibly hide in some statistical properties. Thus, the analysis of
statistical property is the start point of study of this paper. Along this
start point, five equivalent statements of ACO convergence are found in this
paper (see Conlusion 1-5). And these equivalent statements result in the
following conclusion:

ACO may not converges to the optimal solution in practice, but its entropy
is convergent under general condition.

\begin{acknowledgments}
The first author thanks his teacher prof. G.-C. Guo because his main study
methods are learned from his lab. of quantum information. The first author
thanks prof. Z. F. Han's and prof. Z.-W Zhou working at Guo's lab. for they
helping him up till now. The first author thanks prof. J. Zhang, prof. Q. Li
and prof. J. Zhou for their help. The authors thank Dr. Marek Gutowski at
Institute of Physics, Poland for he telling them the careless incorrectness
of one reference. The authors thank prof. walter gutjahr, his encouragement
gave them a great sense of uplift since he is the first man to study the ACO
convergence.
\end{acknowledgments}

\end{document}